\documentclass[times,onecolumn,authoryear]{elsarticle}

\RequirePackage{geometry}
\long\def\@makecaption#1#2{%
    \vskip\abovecaptionskip\footnotesize\bfseries
    \sbox\@tempboxa{#1.~#2}%
    \ifdim \wd\@tempboxa >\hsize
    #1.~#2\par
    \else
    \global \@minipagefalse
    \hb@xt@\hsize{\hfil\box\@tempboxa\hfil}%
    \fi
    \vskip\belowcaptionskip}

\global\bibsep=0pt

\journal{Computers and Electronics in Agriculture}

\usepackage[utf8]{inputenc}
\usepackage{afterpage}
\usepackage{blindtext}
\usepackage{float}
\usepackage{multicol}
\usepackage{subcaption}
\usepackage{url}
\usepackage{amsmath}
\usepackage{slashbox}
\usepackage{booktabs}
\usepackage{multirow}
\usepackage{adjustbox}
\usepackage[normalem]{ulem}
\newcommand{\ra}[1]{\renewcommand{\arraystretch}{#1}}
\usepackage[labelfont=bf, font=small]{caption}

\usepackage[
 anchorcolor=blue,
 colorlinks=true,
 linkcolor=blue,
 urlcolor=black,
 citecolor=blue
]{hyperref}

\newcommand{\comm}[1]{}

\begin{document}

\title{DropLeaf: a precision farming smartphone application for measuring \\pesticide spraying methods}

\author[1]{Bruno Brandoli\corref{cor1}}
\author[3]{Gabriel Spadon\corref{cor1}}
\author[5]{Travis Esau}
\author[5]{Patrick Hennessy}
\author[3]{Andre C. P. L. Carvalho}
\author[4,3]{Jose F. Rodrigues-Jr\corref{cor1}}
\author[4]{Sihem Amer-Yahia}

\cortext[cor1]{Corresponding authors: \texttt{brunobrandoli@gmail.com} (B. Brandoli), \texttt{gabriel@spadon.com.br} (G. Spadon), and \texttt{junio@icmc.usp.br} (J. F. Rodrigues-Jr)}

\address[1]{Dalhousie University -- Halifax, Nova Scotia, Canada}
\address[3]{University of Sao Paulo -- Sao Carlos, SP, Brazil}
\address[4]{CNRS, Univ. Grenoble Alpes -- Grenoble-Alpes, France}
\address[5]{Dalhousie University -- Truro, Nova Scotia, Canada}

\begin{keyword}
    deposition analysis \sep spray coverage characterization \sep water sensitive papers and cards \sep UAVs spray \sep precision farming
\end{keyword}
    
\begin{abstract}
Pesticide application has been heavily used in the cultivation of major crops, contributing to the increase of crop production over the past decades.
However, their appropriate use and calibration of machines rely upon evaluation methodologies that can precisely estimate how well the pesticides' spraying covered the crops. A few strategies have been proposed in former works, yet their elevated costs and low portability do not permit their wide adoption. This work introduces and experimentally assesses a novel tool that functions over a smartphone-based mobile application, named DropLeaf - Spraying Meter. Tests performed using DropLeaf demonstrated that, notwithstanding its versatility, it can estimate the pesticide spraying with high precision. Our methodology is based on image analysis, and the assessment of spraying deposition measures is performed successfully over real and synthetic water-sensitive papers.
The proposed tool can be extensively used by farmers and agronomists furnished with regular smartphones, improving the utilization of pesticides with well-being, ecological, and monetary advantages. DropLeaf can be easily used for spray drift assessment of different methods, including emerging UAV (Unmanned Aerial Vehicle) sprayers.
\end{abstract}

\maketitle
\section{Introduction}
\label{sec:intro}
The total world population is estimated to be 7 billion individuals, with a projection of expanding to 9.2 billion by 2050. This expansion will request almost 70\% more nourishment because of changes in dietary (more dairy and grains) in underdeveloped countries~\citep{FAO2009}. To adapt to such circumstances, it is obligatory to expand the efficiency of the existing cultivation areas, which might be accomplished by a more reliable food chain and the utilization of pesticides \citep{cooperCP2007,kestersonSENSORS2015,wangPeiS2019}. Pesticides are chemical compounds used for decimating weed plants (herbicides), parasitic (fungicides), or insects (insecticides)~\citep{Bon2014}. The utilization of pesticides is spread around the world, representing a 40-billion-dollar yearly budget~\citep{Popp2013} with huge amounts of synthetic compounds (approximately 2 kg per hectare~\citep{Liu2015}) being sprayed over a wide range of harvests to augment the production of food. Current pieces of evidence point that farming is to confront heavier stress from pests, prompting to a higher interest for pesticides~\citep{Popp2013}.

To minimize the risk of crop losses because of herbivorous insects and mites, most of the world's commercial food production systems are subject to several applications of pesticides before being cropped \citep{cooperCP2007,berenstein2018}. In this situation, it is significant that the right measure of pesticide be sprayed on the harvest fields. Excessive amounts of chemicals may leave residues in the produced food alongside ecological tainting \citep{gonzalesrodriguesJCA2008,farhaBC2018,witton2018}. On the other hand, insufficient doses, and there may be areas of the harvest field that are not protected, lessening productivity \citep{dougoudASD2019}. For instance, \cite{boina2012} investigated that spray droplet size is key in the efficacy of pesticides against {\it Asian citrus psyllid}. Meanwhile, irregular spray coverage might cause pest and/or weed resistance, or behavioral avoidance~\citep{PS3773, PS3330}. Many fertilizers are often applied as liquid solutions sprayed into plant leaves and soil \citep{marcalASABE2008}; to assess their pulverization, it is important to quantify the spray coverage, the relative zone secured by the pesticide droplets -- usually composed of the water carrier, active ingredients, and adjutant. In today's precision agriculture, several papers investigate the spray drift from agricultural pesticide sprayers and their consequential economic and environmental effects \citep{preftakesPJ2019}.

The issue of estimating the spray coverage refers to calculating how much pesticide was showered on each piece of the harvest field. The standard way to do that is to disseminate oil- or water-sensitive papers along with the soil and underplant leaves. Then, such cards are covered with a bromoethyl dye that turns blue within the presence of liquid~\citep{Giles2003}. The issue, at that point, progresses towards surveying each card by tallying the number of droplets per unitary area, by drawing their size distribution, and by evaluating the level of the card area that was covered; these measures enable one to gauge the volume of showered pesticide per unitary area of the harvest. If done manually, this procedure is inefficient and may miss some areas. This is when automatic solutions become essential, including the Swath Kit~\citep{Mierzejewski1991}, a pioneer computer-based procedure that utilizes image processing to asses the water-sensitive cards; the USDA-ARS system~\citep{Hoffman2005}, a camera-based framework that uses 1-$cm^2$ samples from the cards to form a pool of sensor information; the DropletScan~\citep{Wolf2003}, a flatbed scanner defined over a proprietary equipment; the DepositScan framework, made of a workstation and a handheld card scanner~\citep{Zhu2011}; and the AgroScan System\footnote{~\url{http://www.agrotec.etc.br/produtos/agroscan/}}, a batch-based outsource Windows-based software that analyzes the collected cards. Every one of these frameworks, though, are inconvenient to convey all through the field, requiring the collection, scanning, and post-processing of the cards, a tedious and labor-intensive procedure.

An option to image capturing systems is to address the characterization of spray application by using wired or remote sensors~\citep{Crowe2005,giles2007}. However, those are costly and necessitate constant maintenance. Very recently, \cite{wangS2019} implemented a novel droplet deposit sensing system based on a sensor to store the deposition. Then, algorithm Q-Learning \cite{watkins1992q} is used to accurately determine the droplet parameters from UAVs (Unmanned Aerial Vehicles). In 2019, \cite{wangPeiS2019} implemented a new capacitor sensor system for measuring the spray deposit of herbicide application. Moreover, several non-imaging spray methods have been developed for spray analysis by means of non-intrusive characterization, such as phase doppler particle analyzers (PDPA) \citep{nuyttens2009}, piezoelectric sensors \citep{hassan2019}, laser diffraction analyzers (e.g., Malvern analyzers \citep{stainerCP2006}), and optical array probes \citep{teske2002}. They were all designed to assess the quality of spray coverage, including the droplets' size and volume.

Alternatively, a number of image-based approaches have been introduced to assess the efficiency of the spraying deposition quality. Such means profit from the advanced innovations found in smartphones \citep{fengMS2015}, which convey computing assets powerful enough to enable a wide scope of uses. In the form of a smartphone application, an image-based system is conceivable as a promptly-accessible tool, portable up to the harvest field, to help ranchers and agronomists estimate the spray coverage, consequently, in decisions concerning where and how to pulverize. This is the present investigation point, wherein we present DropLeaf - Spraying Meter~\footnote{The website can be accessed at \url{http://dropleaf.icmc.usp.br/}}, a wireless application ready to gauge the measure of pesticide showered on water-sensitive papers. DropLeaf enables precision agriculture, with the potential to improve the evaluation of pesticide showering. It utilizes the phone's camera to register pictures of the spray cards and, nearly immediately, it creates evaluations of the spray coverage using methods based on image processing.

In this context, SnapCard was the first pesticide spray coverage tool developed for running over a smartphone~\citep{nansenASD2015,fergusonCOMPAG2016}. Nevertheless, it presents two drawbacks: i) it calculates the coverage area of the water-sensitive paper only, and ii) it does not allow the user to load a photo from the phone's gallery. Dropcard with DropScope is a similar and commercial smartphone application that relies on an external water-card reader; currently, it works under restricted card sizes. Table \ref{tab:apps} compares our proposed solution to the other smartphone applications developed for measuring spraying coverage. It is worth saying that the smartphone application named Gotas \citep{chaim1999} is not covered since it was discontinued; it is no longer distributed. 

\begin{table}[!htbp]
\centering
\resizebox{\textwidth}{!}{\begin{tabular}{|l|l|l|l|} 
\hline
\textbf{Smartphone Application} & \textbf{Cost and Platform} & \textbf{Advantages} & \textbf{Limitations} \\ 
\hline
\begin{tabular}[c]{@{}l@{}}DropLeaf \\ \cite{brandoliSAC2018} \\ \href{http://dropleaf.icmc.usp.br}{Website} \end{tabular}& free, Android  & \begin{tabular}[c]{@{}l@{}}- elaborated user interface \\- it calculates several statistical measures \\- it works with any card size\\- it exports the card measurements\end{tabular} & \begin{tabular}[c]{@{}l@{}}- it runs over Android only \\- the user must load the card previously cropped using an external photo editor\end{tabular} \\ 
\hline
\begin{tabular}[c]{@{}l@{}}SnapCard \\ \cite{nansenASD2015} \\ \href{http://agspsrap31.agric.wa.gov.au/snapcard}{Website} \end{tabular} & free, Android and IOs & \begin{tabular}[c]{@{}l@{}}- elaborated user interface \\- it runs over both Android and IOs platforms \\- it saves the card measurements\end{tabular} & \begin{tabular}[c]{@{}l@{}}- it calculates the coverage area over the water-sensitive paper only \\- it does not allow the users to load card photos from the gallery\end{tabular} \\ 
\hline
\begin{tabular}[c]{@{}l@{}}DropCard with DropScope \\ \href{https://www.sprayx.com.br}{Website}\end{tabular} & commercial, Android & \begin{tabular}[c]{@{}l@{}}- it calculates several statistical measures \\- it saves additional information based on reports\end{tabular} & \begin{tabular}[c]{@{}l@{}}- it demands additional hardware to read the cards \\- the segmentation of bigger drops fails\\- it does not load from the photo gallery, hampering the reproduction of previous analyses \\- it just reads regular size cards (7.6 cm $\times$ 2.6 cm) \\- it runs over Android only\end{tabular}  \\
\hline
\end{tabular}}
\caption{\label{tab:apps}Comparison of different smartphone applications developed for pesticide spraying assessment using water-sensitive paper.}
\end{table}

The remainder of the paper is structured as follows. Section~\ref{sec:approach} describes the steps of the proposed approach to measure the quality of pest control spraying. In addition, it describes the techniques implemented in the mobile application. In Section~\ref{sec:res}, we show the results achieved by our application in comparison to related works. Section \ref{sec:fractal} discusses the use of fractal theory in the task of evaluating spray coverage, and Section~\ref{sec:discussao} reviews relevant aspects related to our results. Conclusions come in Section~\ref{sec:conclusao}.

\section{Methodology}
\label{sec:approach}

In this section, we introduce our methodology, named DropLeaf, to estimate the pesticide spray coverage. The goal is to quantify the coverage area of water-sensitive papers or spray cards, so to help to estimate how adequate was the pesticide pulverization, as discussed in Section~\ref{sec:intro}. DropLeaf builds upon image processing strategies developed on a portable application that is practical on commodity mobile phones. The tool draws three standard measures \citep{Hoffman2005} from the drops observed on the spray cards, producing a numerical summary that allows assessing the spraying:

\begin{itemize}
   \item{{\bf Coverage Area (CA)}: given in percentage of covered area per area unit in $cm^2$;}
   \item{{\bf Volumetric Median Diameter (VMD)}: given by the 50th percentile $D_{V0.5}$ of the diameter distribution;}
   \item{{\bf Relative Span (RS)}: given by $RS=\frac{D_{V0.9}-D_{V0.1}}{D_{V0.5}}$, where $D_{V0.1}$ is the 10th percentile and $D_{V0.9}$ is the 90th percentile of the diameter distribution.}
\end{itemize}

These three measures drive the estimation of the amount of the field covered with pesticide and how well the pesticide was scattered; finer diameters and higher coverage areas indicate a better scattering.

So as to figure out those measures, it is important to gauge the diameter (in micrometers) of each drop observed on a given card. Manually, this is an arduous task that might take hours per card. To mitigate that, DropLeaf utilizes an intricate image processing method that saves time and provides higher accuracy when contrasted to manual examination and previous systems.

\begin{figure*}[!htb]
   \centering
   \includegraphics[width=0.9\linewidth]{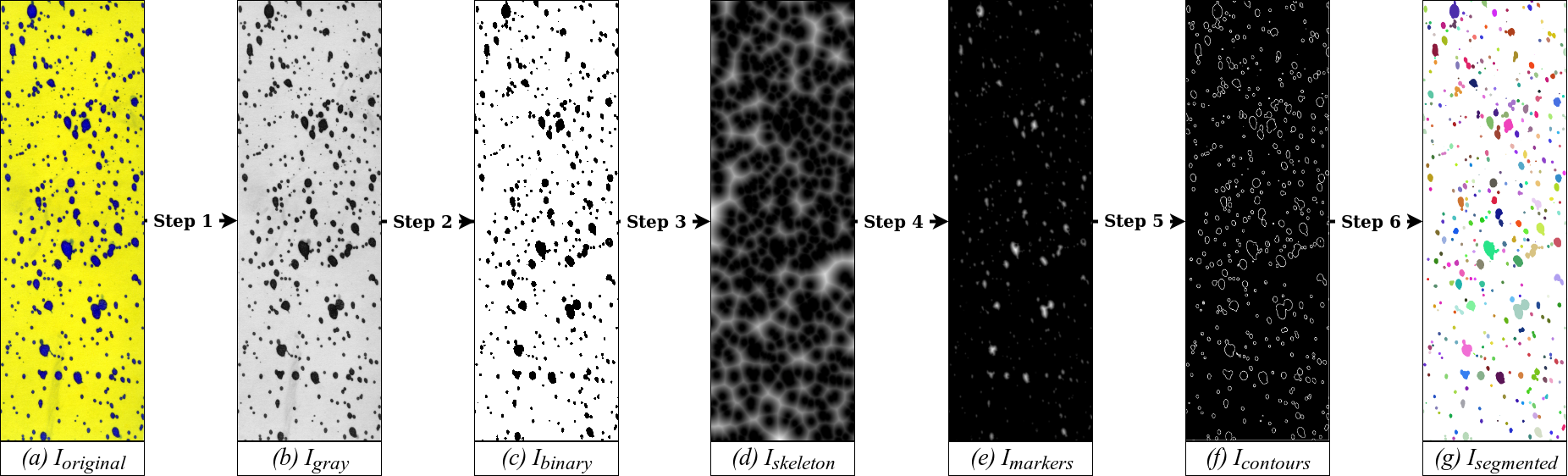}
   \caption
   {The image processing course of DropLeaf. It begins by loading an image of a water-sensitive paper. We then perform a color-space transformation to obtain a grayscale version of the image -- {\bf Step 1}. Subsequently, the grayscale image is binarized to isolate the drops and to remove noise -- {\bf Step 2}. Next, we apply the morphological operation of skeletonization -- {\bf Step 3}, after which we apply a thresholding operation so to emphasize the drops' markers -- {\bf Step 4}. Finally, we use the markers to find the contours of the drops using the marker-based watershed algorithm -- {\bf Step 5}, providing the tool with a well-defined set of droplets -- visualized after {\bf Step 6}.}
   \label{fig:proposta}
\end{figure*}

Figure~\ref{fig:proposta} shows the image processing of DropLeaf, which comprises of six steps carried over a given spray card: (i) color space conversion; (ii) binary noise removal; (iii) morphological operation of skeletonization; (iv) thresholding; (v) identification of droplets via the marker-based watershed algorithm; and (vi) visualization. We clarify each step indicating why it was necessary and how it identifies with the subsequent step. To illustrate the processing steps, we use a running sample whose picture is exhibited in Figure~\ref{fig:proposta}(a).

\subsection{Grayscale transformation}
After the acquisition of an image via the cellphone camera $I_{original}(x,y) = (R_{xy}, G_{xy}, B_{xy}) \in [0,1]^3$, Step 1 converts it to a grayscale image $I_{gray}(x,y) \in [0,1]$.
This is necessary to ease the discrimination of the card surface from the drops that fell on it.
We use the continuous domain of [0,1] so that our formalism can express any color depth; specifically, we use 32 bits for RGB and 8 bits for grayscale.
Color information is not needed as it would make the computation heavier and more complex.
This first step then transforms the image into a grayscale representation, see Figure~\ref{fig:proposta}(b), according to:

\begin{equation}
   I_{gray}(x,y) = 0.299*R_{xy} + 0.587*G_{xy} + 0.114*B_{xy}
\end{equation}

\subsection{Binarization}

Here, the grayscale image $I_{gray}$ passes through a threshold-based binarization process -- Step 2, a usual step for image segmentation.
Since the grayscale is composed of a single color channel, we achieve binarization by choosing a threshold value.
Gray values $I_{gray}(x,y)$ below the threshold become black and white otherwise.
Since spray cards are designed to stress the contrast between the card and the drops, the threshold value is set as a constant value -- we use value $0.35$ corresponding to value $90$ in the 8-bit domain $[0,255]$.
This is a choice that removes noise, and that favors faster processing if compared to more elaborated binarization processes like those based on clustering or on gray-levels distribution.
Figure~\ref{fig:proposta}(c) depicts the result, an image $I_{binary}(x,y) \in \{0,1\}$ given by:

\begin{equation}
   I_{binary}(x,y) =
       \begin{cases}
           0, & \mbox{if }I_{gray}(x,y) < 0.35\\
           1, & \mbox{otherwise}
       \end{cases}
\end{equation}

\subsection{Skeletonization}

At this point, we need to identify marks that will spot each drop individually -- Step 3. We use the morphology operation of distance transform considering the Euclidean norm and a scale factor of 3 \citep{gonzalez2007image}. This algorithm will set the intensity of the white pixels proportional to the distance to their closest black pixel, that is, to the closest drop boundary. The result is a skeleton that emphasizes the inner regions of the drops. Formally, we produce an image $I_{skeleton}$ according to:

\begin{equation}
   I_{skeleton}(x,y) = min(\sqrt{(x-x')^2 + (y-y')^2}),\ \ \forall\ (x',y') \in I_{binary},\ (x',y') = 0
\end{equation}

We normalize the distances computed during the skeletonization using min-max, after which all the pixels' intensities lie within the range [0,1] -- see Figure~\ref{fig:proposta}(d).

\subsection{Thresholding}
Next, we refine the skeleton image to mark the positions of the drops properly. We use a filter based on a threshold value -- we empirically defined the value of 0.17, but the user can interactively redefine it according to the number and structure of drops. After that, only the strongest most central pixels of each drop will remain in the image, as illustrated in Figure \ref{fig:proposta}(e)

\begin{equation}
   I_{markers}(x,y) =
       \begin{cases}
           0, & \mbox{if }I_{skeleton}(x,y) < 0.17\\
           (x,y), & \mbox{otherwise}
       \end{cases}
\end{equation}

\subsection{Marker-based watershed segmentation}

In the last step -- Step 4, with drops properly marked on the image, we proceed to the drop identification considering the previously identified contours.
To this end, we used the marker-based watershed segmentation.
Watershed~\citep{vincent1991watersheds} is a technique that considers an image as a topographic relief in which the gray level of the pixels corresponds to their altitude.
The transform proceeds by simulating the flooding of a landscape starting at the local minima.
This process forms basins that are gradually fulfilled with water.
Eventually, the water from different basins meet, indicating the presence of ridges (boundaries); this is an indication that a segment was found and delimited.
The process ends when the water reaches the highest level in the color-encoding space.
The problem with the classical watershed is that it might over-segment the image in the case of an excessive number of minima.
For better precision, we use the marker-controlled variation of the algorithm proposed by \citep{gaetano2012marker}.
This variation is meant for images in which the shapes are previously marked.
Given the markers, the marker-based watershed proceeds by considering as minima only the pixels within the markers' boundaries.
Watershed is an iterative algorithm computationally represented by a function {\it watershed(Image i, Image[] markers)}.
We use such a function to produce a set of segments (drops) over the gray-level image $I_{gray}$ while considering the set of markers identified in the image $I_{markers}$, as follows:
\begin{equation}
   contours[] = watershed(I_{gray},findContours(I_{markers}))
\end{equation}

\noindent
where $Image[]~findContours(Image~i)$ is a function that, given an image, returns a set of sub-images corresponding to the markers; watershed is a function that, given an input image and a set of markers corresponding to subsets of pixels, produces a set of segments stored in an array of contours, which we illustrate in Figure~\ref{fig:proposta}(f).

We use the product of watershed to produce our final output $I_{segmented}$ by drawing the segments over the original image, as illustrated in Figure~\ref{fig:proposta}(g).
Notice that the last image, $I_{segmented}$, is meant only for visualization.
The statistical analysis over the drops' shapes is computed over the set of segments.

\subsection{Diameter processing}

After the segmentation, we have a set of segments, each corresponding to a drop of pesticide.
The final step is to compute the measures presented at the beginning of this section: coverage area (CA), volumetric median diameter (VMD), and relative span (RS).
Since we have the segments computationally represented by an array of binary matrices, we can calculate the area and the diameters of each drop by counting each matrix's pixels.
After counting, it is necessary to convert the diameter given in pixels into a diameter given in micrometers ($\mu m$), which, for the i-th drop, goes as follows:
\begin{equation}
   diameter_{\mu m}(drop_i) = width_{px}(drop_i)*\frac{width^{card}_{\mu m}}{width^{card}_{px}}
   \label{eq:diametermu}
\end{equation}

\noindent
where, $width_{px}(drop_i)$ is the width in pixels of the $i$-th drop; $width^{card}_{px}$ is the width of the card in pixels; and $width^{card}_{\mu m}$ is the width of the card in micrometers.
Notice that we used $width$, but we could have used $height$ as well; what matters is that the fraction provides a conversion ratio given in $px/\mu m$, which is not sensitive to the axis; horizontal or vertical, the ratio is the same for a non-distorted image.

Notice that we obtain $width_{px}(drop_i)$ and $width^{card}_{px}$ via image processing, after the segmentation method; meanwhile, $width^{card}_{\mu m}$ is a constant provided by the user, corresponding to the real-world width of the card.
Also, notice that we are considering that the diameter corresponds to the horizontal axis (the width) of the drop; it is possible, however, that the diameter corresponds to the vertical axis, in which case the formulation is straightly similar.
Choosing between the horizontal and the vertical axes might be tricky in case the drop is elliptical, rather than circular.
We solved this issue by extracting the diameter from the area of the drop.
We use the formula of the circle area $a_{circle}=\pi*radius^2 = \pi*(\frac{diameter}{2})^2$.
With simple algebra, we conclude that given the area in pixels of the $i$-th drop, its diameter in pixels is given by the following equation:
\begin{equation}
   diameter_{px}(drop_i) = 2*\sqrt{\frac{area_{px}(drop_i)}{\pi}}
   \label{eq:diameterrefined}
\end{equation}

Rewriting Equation~\ref{eq:diametermu} by means of Equation~\ref{eq:diameterrefined}, we get:
\begin{equation}
   diameter_{\mu m}(drop_i) = 2*\sqrt{\frac{area_{px}(drop_i)}{\pi}}*\frac{width^{card}_{\mu m}}{width^{card}_{px}}
   \label{eq:diametermufinal}
\end{equation}

Once the diameter is converted into micrometers, it becomes trivial to compute all the measures that support the spray card analysis, as described at the beginning of Section~\ref{sec:approach}.

\subsection{Implementation}

The use of mobile devices to perform automatic tasks has increased fast~\citep{fengMS2015}. The reasons are the recent advances in hardware, such as sensors, processors, memories, and cameras. Thereby, smartphones have become platforms for applications of image processing and computer vision~\citep{farinellaPR2015,brandoliCEA2016}.

\begin{figure*}[!htb]
   \centering
   \includegraphics[width=.9\linewidth]{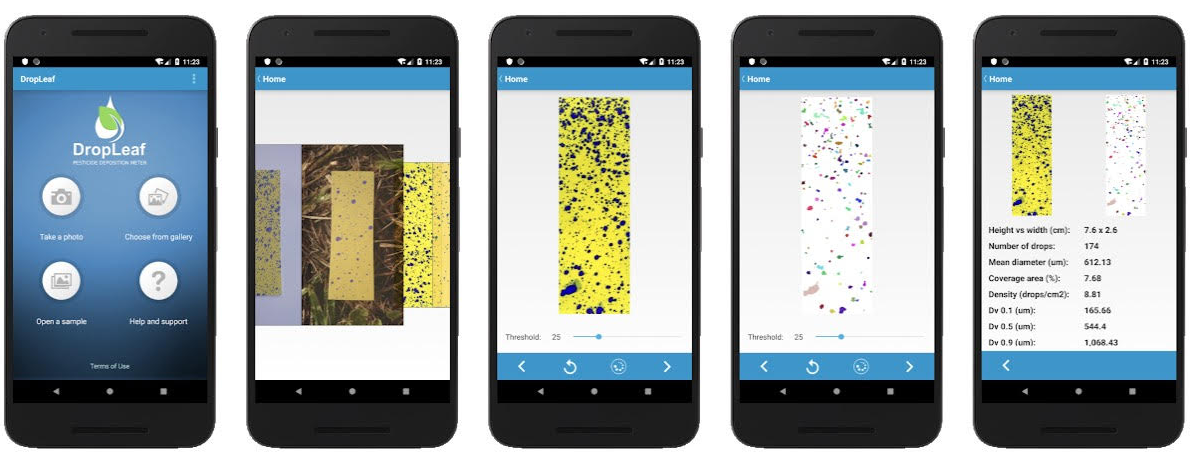}
   \caption{Screenshots of our fully-functional app. From left to right: home screen; photo capture of the water-sensitive-card; segmentation process; segmentation result; and computed metrics. Freely available for download on \href{https://play.google.com/store/apps/details?id=upvision.dropleaf}{GooglePlay.}}
   \label{fig:sample}
\end{figure*}

Mobile devices are adequate to perform real-time tasks {\it in situ}, far from the laboratory.
In this context, besides the methodology, this work's contribution is the development of a mobile application to measure the quality of pesticide spraying on water-sensitive cards. For implementation, we used methods from the OpenCV library\footnote{~\url{http://opencv.org}}, and Java was the programming language.
The application is fully functional, as depicted in Figure~\ref{fig:sample}, and available in the Google Play platform at \url{https://play.google.com/store/apps/details?id=upvision.dropleaf}.

\begin{table*}[hbt]
   \centering
   \ra{1.3}
   \begin{tabular}{ccccc}
       \toprule
       \multicolumn{5}{c}{\textbf{Dropleaf}} \\ 
       \midrule
       \multicolumn{2}{c}{Diameter ($\mu m$)}  &   \multirow{2}{*}{Area ($\mu m^2$)} &  \multirow{2}{*}{Density (drop/$cm^2$)} &  \multirow{2}{*}{Coverage Area (\%)} \\ 
       \cmidrule{1-2}
       Controlled & Measured &  &  &  \\ 
       \midrule
       50         & 58       & 2,693            & 594.31        & 5.32\%  \\ 
       100        & 141      & 15,687           & 399.01        & 15.01\% \\ 
       250        & 246      & 53,470           & 229.73        & 23.46\% \\ 
       500        & 467      & 214,970          & 37.20         & 11.8\%  \\ 
       1,000      & 1,009    & 901,811          & 3.65          & 3.72\% \\
       \bottomrule
   \end{tabular}
   \caption{DropLeaf drop identification over the control card by enterprise Hoechst.}
   \label{tbl:Results}
\end{table*}

\begin{table}
\centering
\ra{1.5}
\begin{adjustbox}{width=\textwidth}
\begin{tabular}{cccccccccccc}
 \toprule
 \multirow{2}{*}{Diameter ($\mu m$)} & \multicolumn{3}{c}{\textbf{DropLeaf}} & & \multicolumn{3}{c}{\textbf{DepositScan}} & & \multicolumn{3}{c}{\textbf{Microscope}}\\ 
 \cmidrule{2-4} \cmidrule{6-8} \cmidrule{10-12}
 {} & Area ($\mu m ^2$) & Diameter ($\mu m$) & Error && Area ($\mu m ^2$) & Diameter ($\mu m$) & Error && Area ($\mu m ^2$) & Diameter ($\mu m$) & Error \\ 
\midrule
    50    & 2,693   & 58    & 16\%  && 6,093   & 88  & 76\%  && 3,390   & 66    & 32\%  \\ 
    100   & 15,687  & 141   & 41\%  && 21,505  & 165 & 65\%  && 15,906  & 142   & 42\%  \\ 
    250   & 53,470  & 246   & 1.6\% && 52,688  & 259 & 3.6\% && 45,342  & 240   & 4\%   \\ 
    500   & 214,970 & 467   & 6.6\% && 196,236 & 500 & 0\%   && 201,924 & 507   & 1.4\% \\ 
    1,000 & 901,811 & 1,009 & 0.9\% && 777,954 & 995 & 0.5\% && 797,752 & 1,008 & 0.8\% \\ 
\bottomrule
\end{tabular}
\end{adjustbox}
\caption{DropLeaf compared to tool DepositScan and a stereoscopic microscope with respect to the control card.}
\label{tab:microscope}
\end{table}

\begin{figure}[!b]
   \centering
   \setlength{\fboxsep}{0pt}
   \fbox{\includegraphics[width=.2\linewidth]{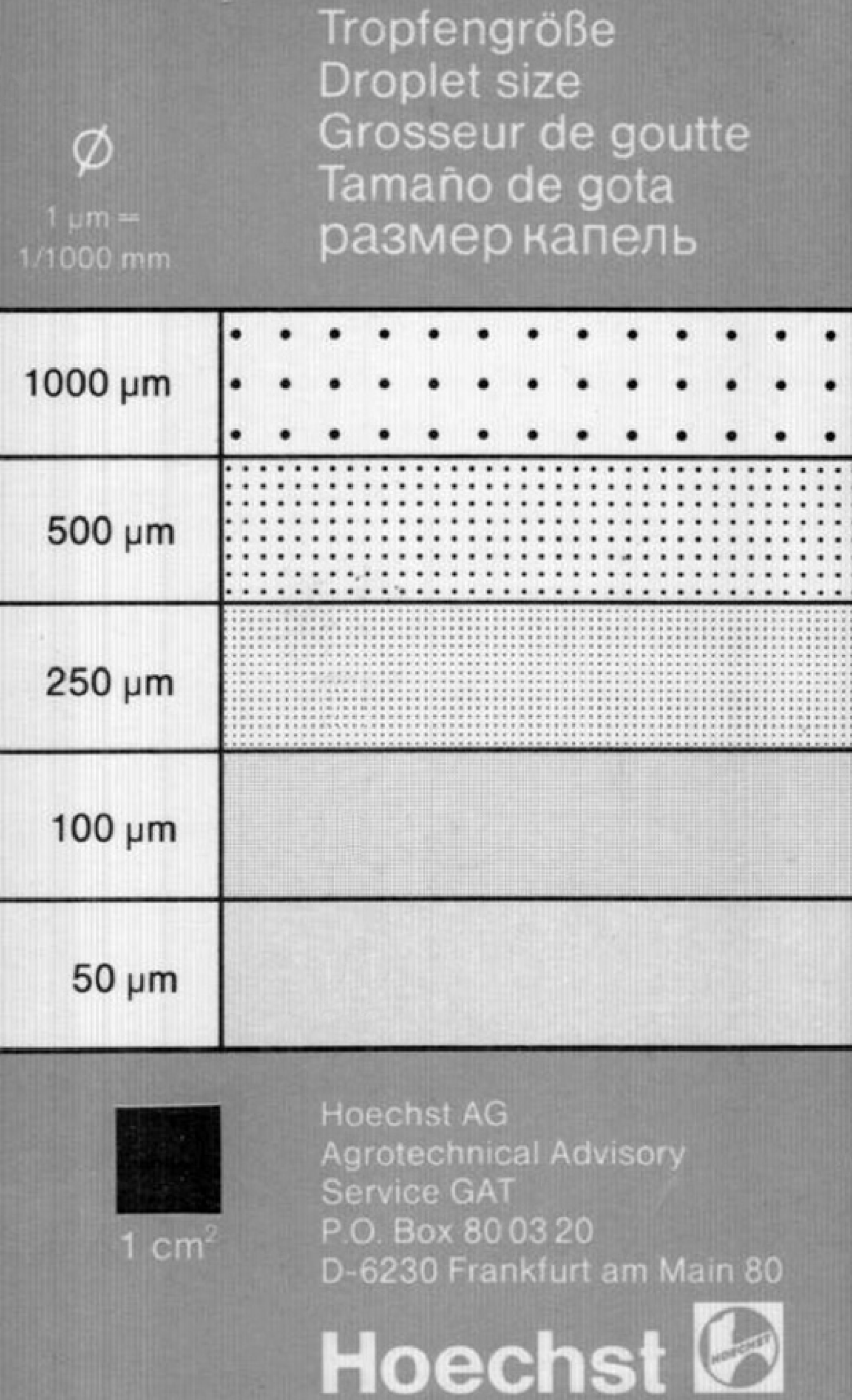}}
   \caption{Control card provided by {\it Hoechst}.}
   \label{fig:controcard}
\end{figure}

\afterpage{

\begin{figure*}[!htb]
   \centering
   \setlength{\fboxsep}{0pt}%
   \begin{subfigure}{.165\textwidth}
     \begin{subfigure}{.5\textwidth}
       \centering
       \fbox{\includegraphics[width=.98\linewidth]{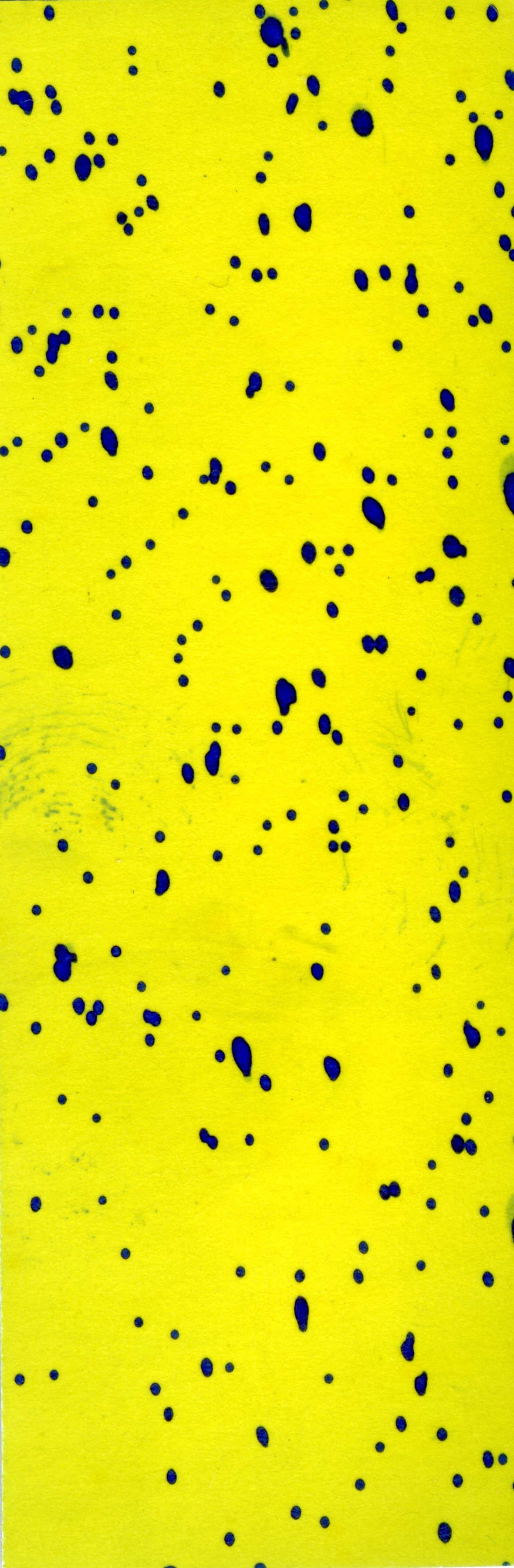}}
     \end{subfigure}%
     \begin{subfigure}{.5\textwidth}
       \centering
       \fbox{\includegraphics[width=.98\linewidth]{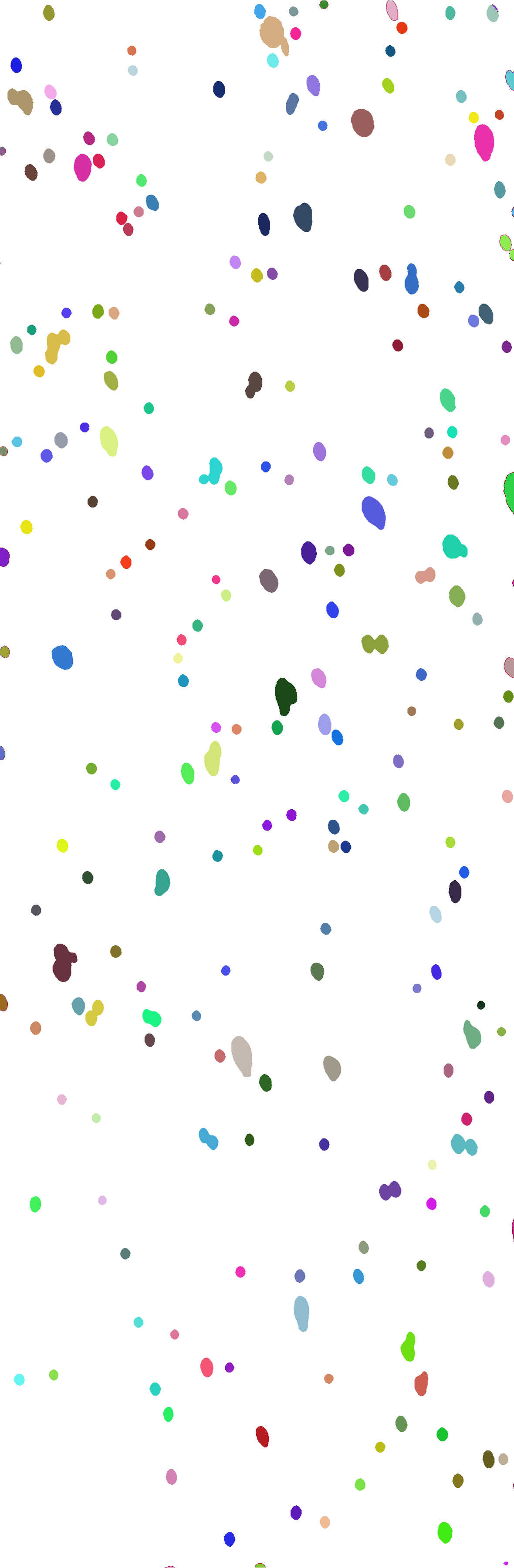}}
     \end{subfigure}%
     \caption{}
     \label{fig:sparse1-7}
   \end{subfigure}
   \hspace{-1.8mm}
   \begin{subfigure}{.165\textwidth}
     \begin{subfigure}{.5\textwidth}
       \centering
       \fbox{\includegraphics[width=.98\linewidth]{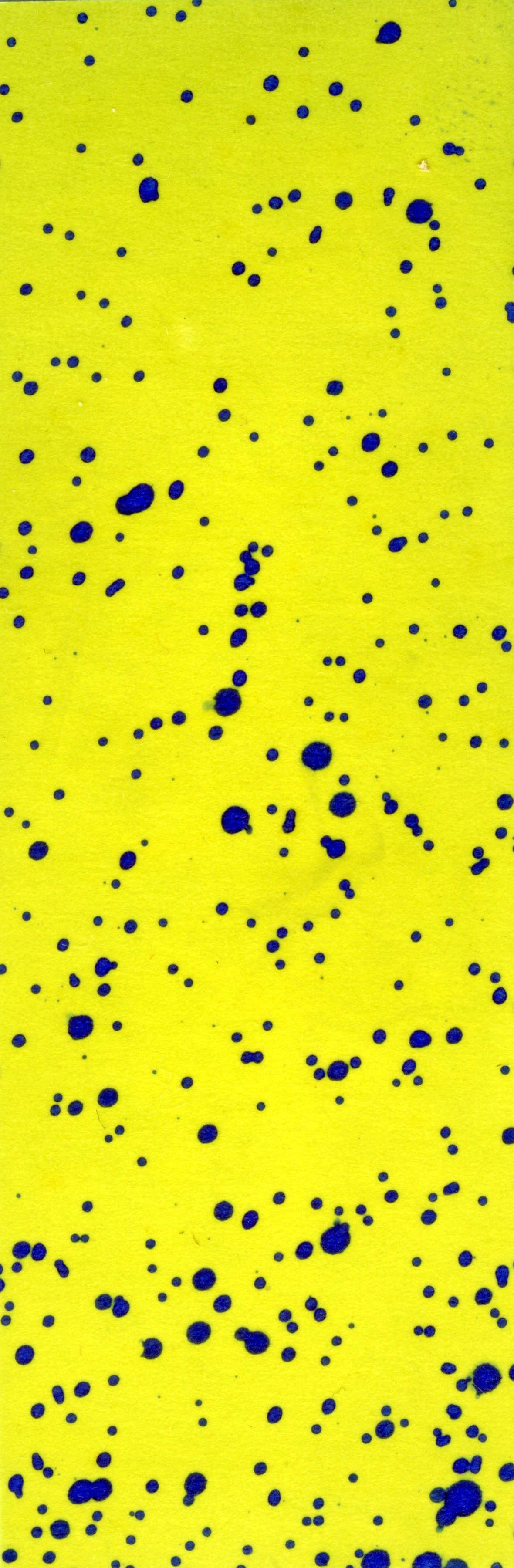}}
     \end{subfigure}%
     \begin{subfigure}{.5\textwidth}
       \centering
       \fbox{\includegraphics[width=.98\linewidth]{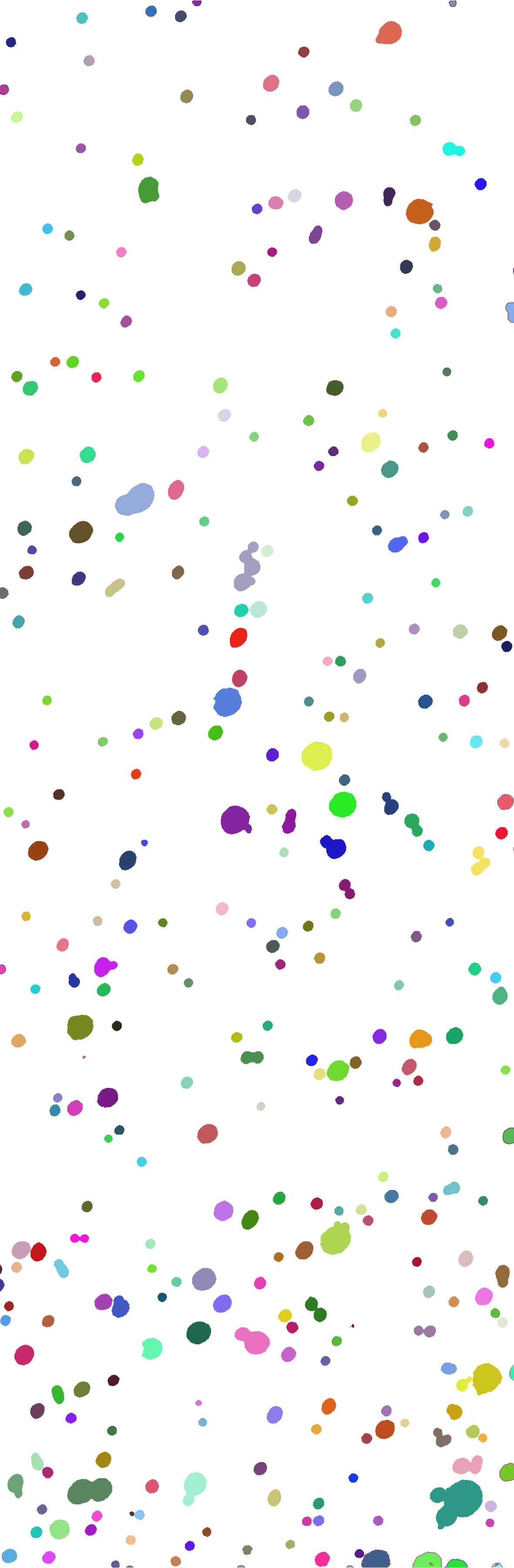}}
     \end{subfigure}%
     \caption{}
     \label{fig:sparse3-6}
   \end{subfigure}%
   \hfill
   \begin{subfigure}{.165\textwidth}
     \begin{subfigure}{.5\textwidth}
       \centering
       \fbox{\includegraphics[width=.98\linewidth]{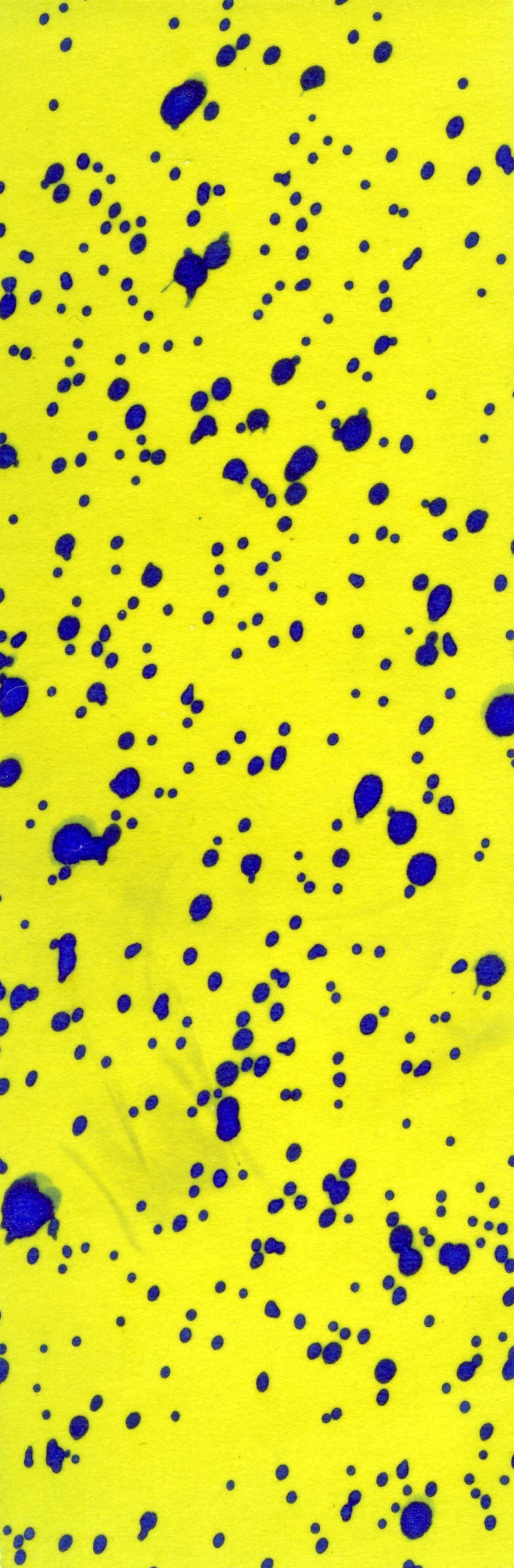}}
     \end{subfigure}%
     \begin{subfigure}{.5\textwidth}
       \centering
       \fbox{\includegraphics[width=.98\linewidth]{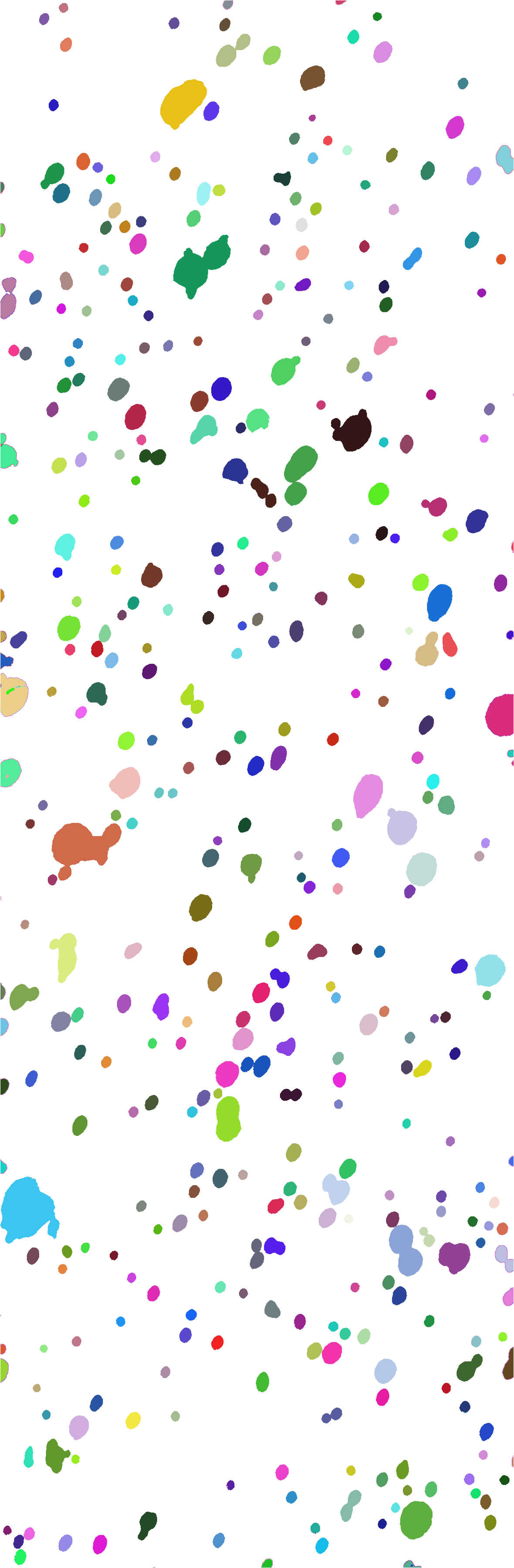}}
     \end{subfigure}%
     \caption{}
     \label{fig:medium1-5}
   \end{subfigure}%
   \hfill
   \begin{subfigure}{.165\textwidth}
     \begin{subfigure}{.5\textwidth}
       \centering
       \fbox{\includegraphics[width=.98\linewidth]{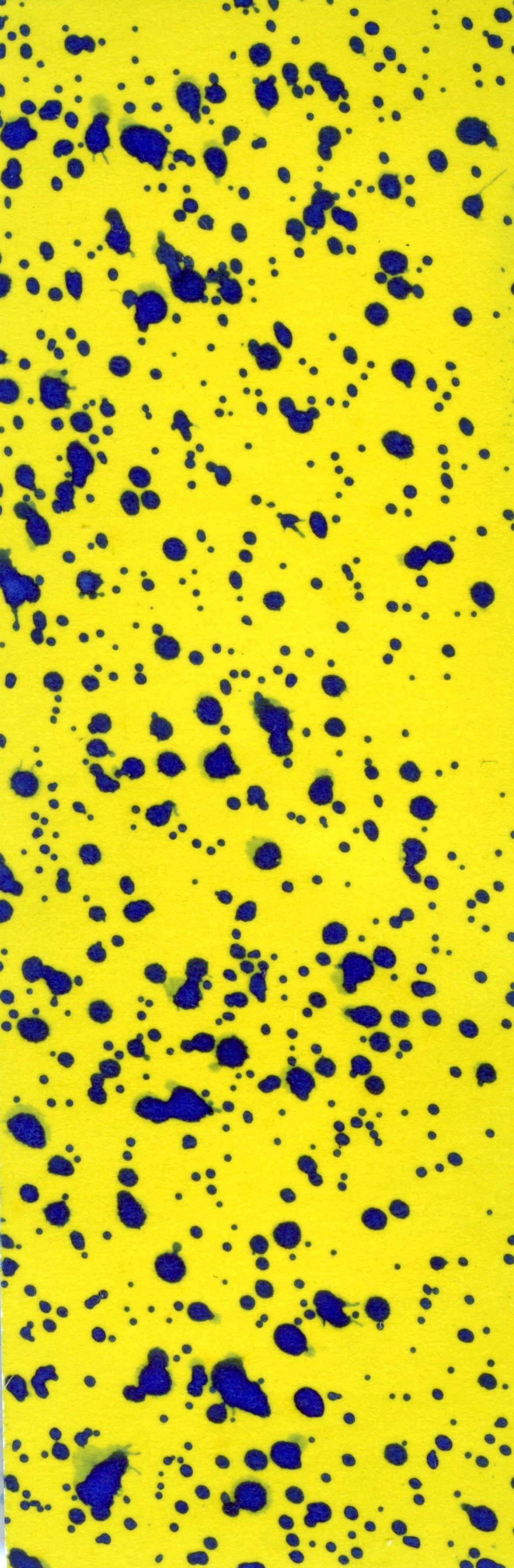}}
     \end{subfigure}%
     \begin{subfigure}{.5\textwidth}
       \centering
       \fbox{\includegraphics[width=.98\linewidth]{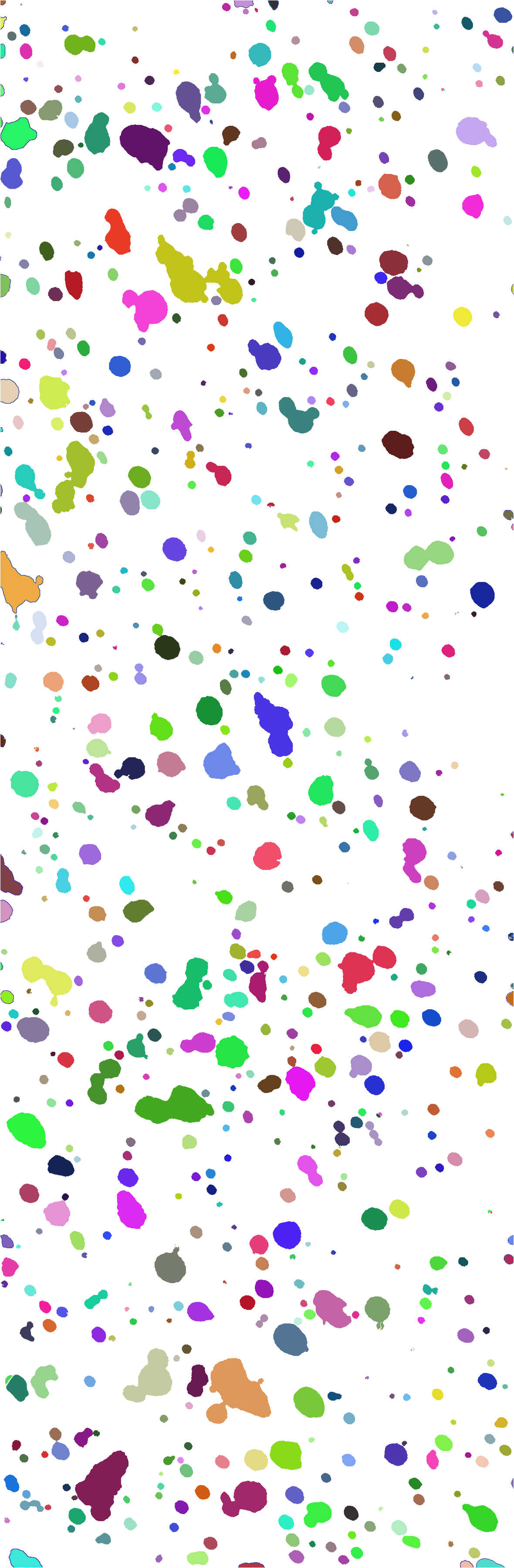}}
     \end{subfigure}%
     \caption{}
     \label{fig:medium3-2}
   \end{subfigure}%
   \hfill
   \begin{subfigure}{.165\textwidth}
     \begin{subfigure}{.5\textwidth}
       \centering
       \fbox{\includegraphics[width=.98\linewidth]{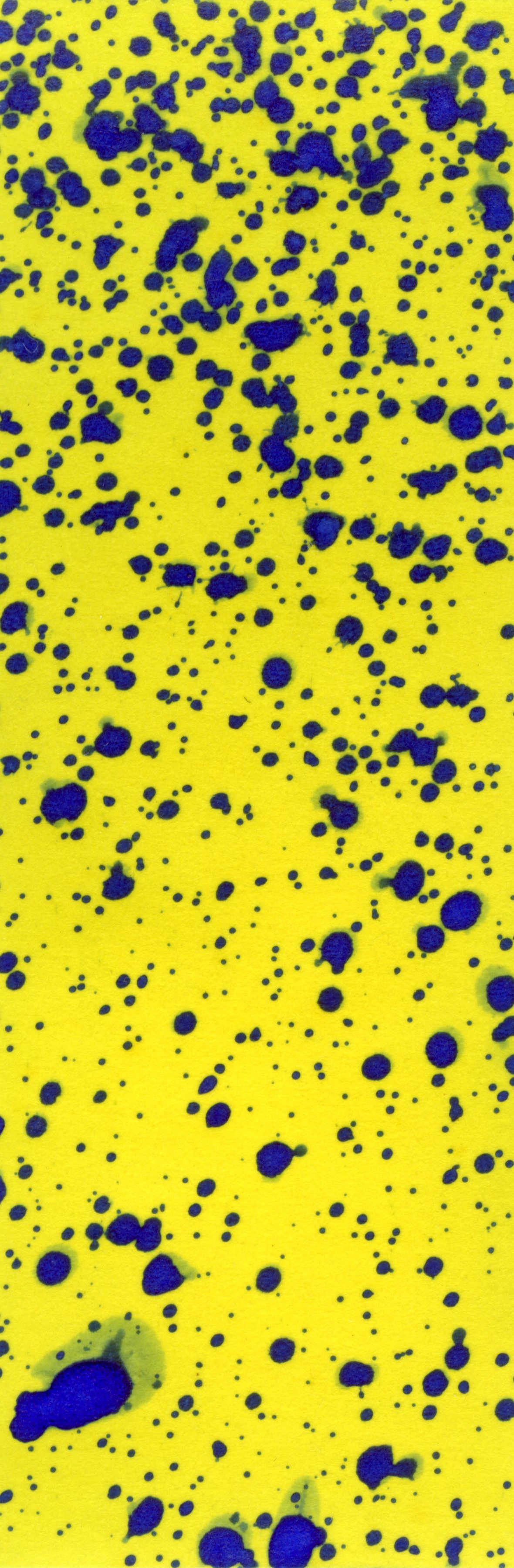}}
     \end{subfigure}%
     \begin{subfigure}{.5\textwidth}
       \centering
       \fbox{\includegraphics[width=.98\linewidth]{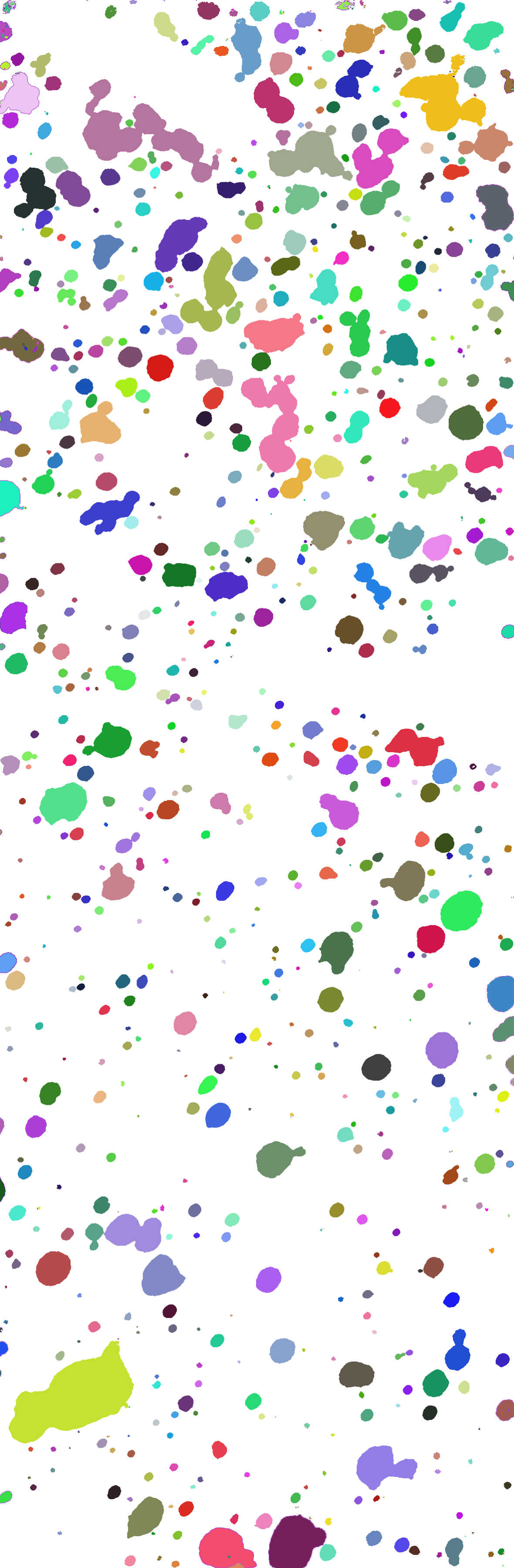}}
     \end{subfigure}%
     \caption{}
     \label{fig:dense1-8}
   \end{subfigure}%
   \hfill
   \begin{subfigure}{.165\textwidth}
     \begin{subfigure}{.5\textwidth}
       \centering
       \fbox{\includegraphics[width=.98\linewidth]{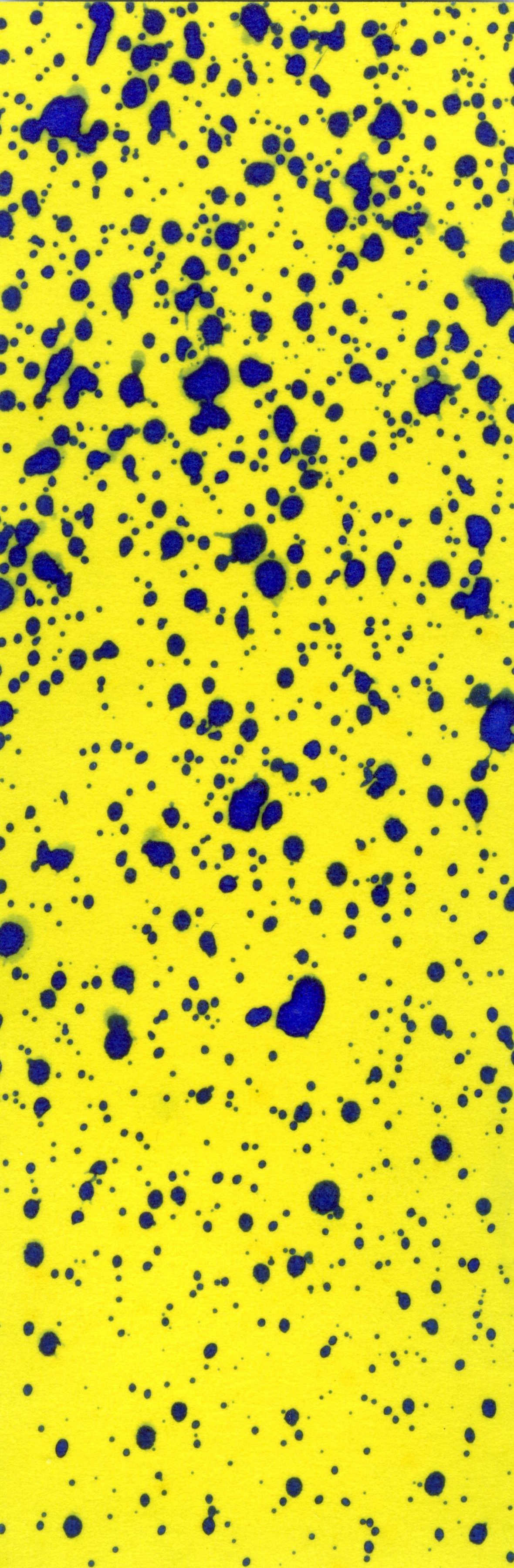}}
     \end{subfigure}%
     \begin{subfigure}{.5\textwidth}
       \centering
       \fbox{\includegraphics[width=.98\linewidth]{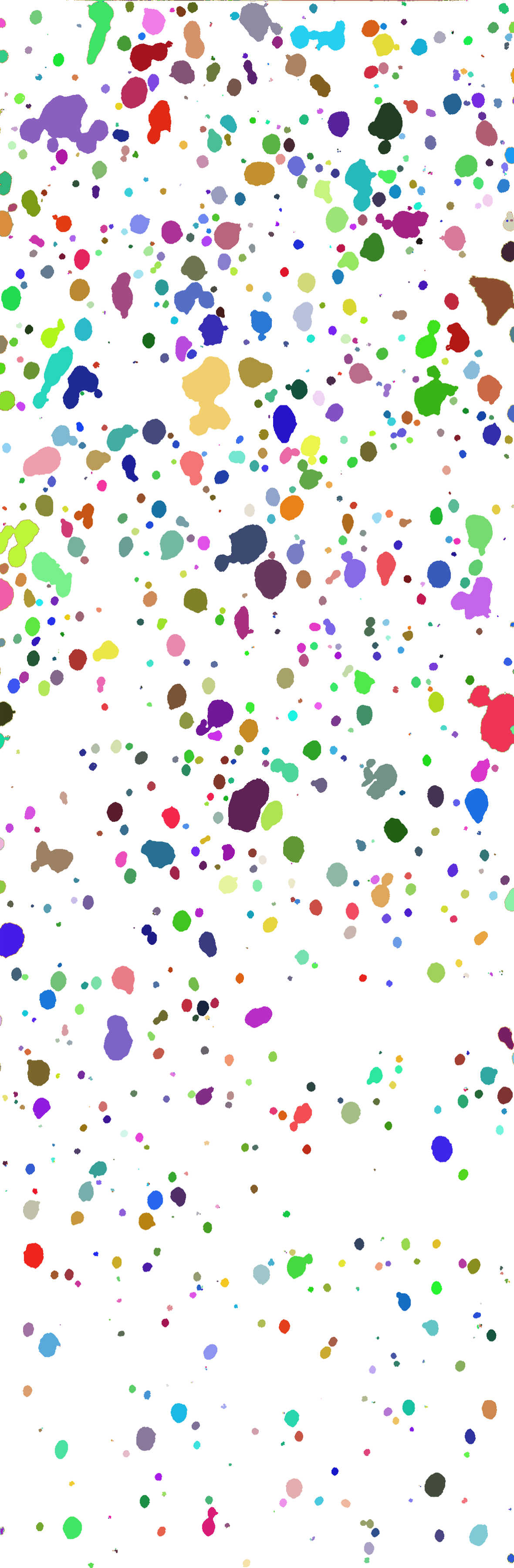}}
     \end{subfigure}%
     \caption{}
     \label{fig:dense2-3}
   \end{subfigure}%
   \caption{Drop identification over cards used in a real crop.
   We categorized the cards as: sparse -- images (a) and (b); medium -- images (c) and (d); and dense -- images (e) and (f).
   The ones on the left are the original cards, and at the right are the segmented cards.}
   \label{fig:realcards}
\end{figure*}

\begin{table*}[htb]
   \centering
   \ra{1.3}
   \begin{tabular}{rcccccc}
       \toprule
       \multicolumn{7}{c}{\textbf{Dropleaf}} \\ 
       \midrule
       \multirow{2}{*}{Sample} & \multirow{2}{*}{Drops} & \multirow{2}{*}{Area ($\mu m^2$)} & Density       & Coverage     & Volumetric Median  & Relative \\
            &                        &                                   &      (drop/$cm^2$) & Density (\%) & Diameter ($\mu m$) & Span    \\ 
       \midrule
       sparse (a) & 255                    & 250,138                           & 12.90         & 4.54      & 452                & 1.22              \\ 
       sparse (b) & 359                    & 261,464                           & 18.16         & 6.45       & 425                & 1.55              \\ 
       medium (c) & 448                    & 355,712                           & 22.67         & 9.99      & 448                & 1.83              \\ 
       medium (d) & 444                    & 357,005                           & 22.46         & 9.71    & 428                & 2.22              \\ 
       dense (e)  & 923                    & 364,749                           & 46.71         & 18.22      & 246                & 3.75              \\ 
       dense (f)  & 1,150                  & 215,090                           & 58.19         & 15.44     & 239                & 3.40              \\ 
       \bottomrule
   \end{tabular}
   \caption{Drop assessment over cards used in a real crop.}
   \label{tbl:DropleafResults}
\end{table*}
}

\section{Experimental results}
\label{sec:res}

In this section, we evaluate our methodology in measuring the spray coverage deposition.
The goal is to correctly measure the spray drops in terms of density (percentage of coverage per $cm^2$) and drop diameter.
The first set of experiments was conducted over a control card used by enterprise {\it Hoechst}, demonstrating the accuracy in controlled conditions.
The second set of experiments was conducted over a real water-sensitive card used on soy crops, demonstrating that the application works even during {\it in situ} conditions.

\subsection{Control-card experiments}
We use the control card provided by the {\it Agrotechnical Advisory} of German enterprise {\it Hoechst}.
The card holds synthetic drops with sizes 50$\mu m$, 100$\mu m$, 250$\mu m$, 500$\mu m$, and 1,000$\mu m$, as shown in Figure~\ref{fig:controcard}. It is used to calibrate equipment and to assess the accuracy of manual and automatic measuring techniques.
Since the number and sizes of drops are known, this first experiment works as a controlled validation.

To measure the drops, we used a smartphone to capture the image of the card. In Table~\ref{tbl:Results}, we present the average diameter of the drops, the area covered by the drops given in $cm^2$, the density given in drops per $cm^2$, the coverage area given in percentage of the card area, and the volumetric median diameter.
We do not present the volumetric median and the relative span because, as all the drops are equal, these values become not significant.
From the table, we conclude that the methodology's accuracy is in accordance with the controlled protocol; that is, the known and measured diameters match in most cases.
Notice that it is not possible to achieve a perfect identification because of printing imperfections and numerical issues that inevitably arise at the micrometer scale.
For example, for 1,000 $\mu m$ drops, the average diameter was 1,009 $\mu m$.
This first validation was necessary to test the tool's ability in telling apart card background and drops.

Still using the control card, Table~\ref{tab:microscope} compares the coverage area and the average diameter measured by DropLeaf, by tool DepositScan, and by a stereoscopic microscope (provided in the work of \citep{Zhu2011}.
The results demonstrated that the stereoscopic microscope had the best performance as expected since it is a fine-detail laborious inspection.
DropLeaf presented the best results after the microscope, beating the precision of DepositScan for all the drop sizes, but 500 $\mu m$; for 1,000 $\mu m$ drops, the two tools had a similar performance, diverging in less than $1\%$.
In the experiments, one can notice that the bigger the drop, the smaller the error, which ranged from 41\% to less than 1\%.
For bigger drops, the drop identification is next to perfect. 
When measuring drops as small as 50 $\mu m$, a single extra pixel detected by the camera is enough to produce a big error. This problem was also observed in the work of \citep{Zhu2011}.

By analyzing the data, we concluded that the error due to the size scale is predictable.
Since it varies with the drop size, it is not linear; nevertheless, it is a pattern that can be corrected with the following general equation:

\begin{equation}
diameter' = a*diameter^b
\end{equation}

In the case of our tool, we used $a=0.2192733$ and $b=1.227941$.
These values shall vary from method to method, as we observed for DepositScan and the stereoscopic microscope.

\subsection{Card experiments}
\label{sec:production}

In the second set of experiments, we used six cards evaluated in the work of \citep{cunhaBE2012}. Similar to them, we categorized the cards into three groups of two cards, which we classified as sparse, medium, and dense with respect to the density of drops, as seen in Figure~\ref{fig:realcards}.
These experiments tested the methodology's robustness, its ability to identify drops even when they are irregular and/or they have touching borders.
Table~\ref{tbl:DropleafResults} shows the numerical results, including the number of drops, the coverage area, the density, the coverage area, the volumetric median diameter, and the relative span. It is necessary to interpret the table along with Figure~\ref{fig:realcards}, which presents the drops as identified by our methodology.
In the figure, it is possible to inspect the four first measures visually.
It is also possible to see that the right-hand side images (the tool's results stressed with colored drops) demonstrate that the segmentation matches the expectations of a visual inspection; the drops at the left are perfectly identified on the right.
Other features are also noticeable; density, for instance, raises as we visually inspect Figure~\ref{fig:realcards}(a) through Figure~\ref{fig:realcards}(f); the corresponding numbers in the table raise similarly.
Counting the number of drops requires close attention and much time; for the less dense Figures \ref{fig:realcards}(a) and \ref{fig:realcards}(b), however, one can verify the accuracy of the counting and segmentation.

The last two measures, VMD and RS, provide parameters to understand the drops' diameters' distribution. For example, one can see that being denser, cards (e), and (f) had a smaller median and a larger diameters span. These measures indicate that the spraying is irregular and that it needs adjustment. Meanwhile, cards (a) and (b) are more regular, but not as dense as desired, with many blank spots. Cards (c) and (d), in turn, have more uniform spraying and more regular coverage.

\section{Fractal analysis}
\label{sec:fractal}
In this section, we present early experiments related to using fractal theory to express the spraying pattern of droplets on a water-sensitive paper.

\subsection*{Fractal theory}
Fractal geometry provides a mathematical model for complex objects found in nature. In contrast to the Euclidean geometry, the fractal dimension assumes that an object might have a non-integer dimensionality. Estimating an object's fractal dimension is essentially related to its complexity, which can be measured in terms of how it occupies the space. For instance, the fractal dimension has been applied in texture analysis and shape measurement \citep{brandoliPDE2013}, among other applications. Although there are different methods for calculating the fractal dimension, the box-counting method is the most frequently used for measurements in various application fields -- specifically, the spray-card problem is particularly adherent to the box-counting method. Its procedure is as follows:

\begin{equation}
	D = 2 - \lim_{\sigma \to 0} \frac{log \, N(\sigma)}{log \, (1/\sigma)}
\end{equation}

\noindent where $N(\sigma)$ is the least number of boxes of length $\sigma$ to completely cover the object, scaled down by a ratio of $1/\sigma$. Given a binary image of $M \times M$ pixels, where $M$ is a power of 2, first generate a set of box sizes $\sigma$ for laying grids on the image. Subsequently, each grid becomes a box of size $\sigma \times \sigma$, and the number of boxes $N(\sigma)$ needed to cover the object is counted completely. Finally, the limit is calculated using the linear regression of the curve $\log 1/\sigma \times \log N(\sigma)$. The fractal dimension is computed by 
$D = 2 - |\alpha|$, where $\alpha$ is the slope of the estimated line.

\subsection*{Spraying pattern analysis}
\begin{figure}[!htb]
   \centering
   \includegraphics[width=0.5\linewidth]{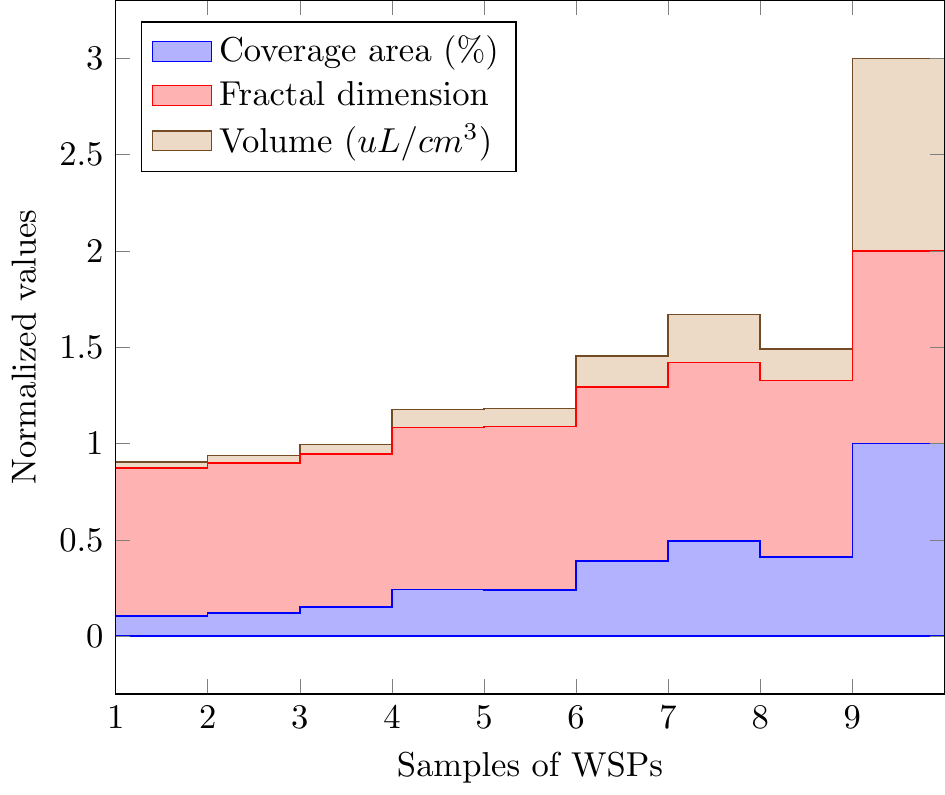}
   \caption{The plot of nine samples of water-sensitive paper. One can observe the correlation among measures of coverage area, volume, and fractal dimension. The x-axis corresponds to the number of the sample; the y-axis corresponds to the normalized output of values of the three measures.}
   \label{fig:correlation}
\end{figure}

In this experiment, we analyzed images of water-sensitive paper by means of fractal dimension. Our goal was to find pieces of evidence that the space occupation of droplets on a water-sensitive paper has a straight relationship to its fractal dimensionality. In Figure \ref{fig:correlation}, we present experiments carried over nine different real samples. In the figure, one can see that the fractal dimension is highly correlated to quantitative spraying measures of coverage area (\%) and volume ($uL/m^3$). The conclusion is straight: the higher the fractal dimension's value, the higher the coverage area, and the volume of sprayed pesticide. This early conclusion allows us to speculate about using the fractal dimension as one quantitative measure to describe droplets' regularity over the water-sensitive paper.

\section{Discussion of Results}
\label{sec:discussao}
This section examines issues as when creating advances for spray card investigation. We confronted such issues during our work; here, we examine them as a further contribution to aid scientists dealing with the same or related aspects.

\subsection{Coverage factor}
\label{subs:coverage}
After our experiments, we observed that when the spraying gets excessively thick, it is not possible to properly detect the drops, regardless of which system is utilized for estimation; data about the number of drops, and their diameter distribution, cannot be cast any longer. This impact was brought up by \cite{Fox2003}, who claims that an absolute coverage of the card above 20\% leads to questionable results; and a coverage near 70\% is unfeasible. 

This is due to the fact that, with an excessive amount of spraying, the drops fall excessively close, causing overlaps; visually, it is like two or more drops become one. This phenomenon occurs because of the water drops' intermolecular attractions, which makes them combine, shaping greater drops. 

As a result, caution is required, regardless of which procedure of assessment, whenever the absolute coverage region sums up over 20\%, a situation when the measurements lose accuracy, and one can depend only on the coverage area for basic decision making. Despite the fact that the diameter is not measurable, the enormous drops that may be observed demonstrate an intemperate measure of pesticide and/or a malfunctioning of the spray device.

\subsection{Angle of image capture}
\label{subs:angle}
We likewise observed that, for all the research done so far, including ours, the image processing technique used to identify the drops works only if the card photo's capture angle is 90 degrees. That is, the viewing angle of the camera/scanner must be orthogonal to the spray card surface. This is important in light of the fact that the pixels of the picture are converted into real-world dimensions to express the diameter of the drops in $\mu m$; that is, the components of the picture must be homogeneous concerning scale. In the event that the capture angle is not proper, the picture becomes misshaped, producing different scales in each region of the picture. For flatbed scanners, this is direct to ensure; notwithstanding, for handheld gadgets (cameras and cell phones), extra care is required. In such cases, one may require a special protocol to capture the image, such as utilizing a tripod. This issue may likewise be solved by methods of image processing, which demand extra research and experimentation.

\subsection{Minimum dots per inch (dpi)}
\label{subs:dpi}
Our trials also showed that there must be a minimum amount of data on the spray card pictures to accomplish the ideal accuracy with respect to the drops' diameters. This minimum is expressed by the {\it dots per inch} (dpi) property of the imaging process; dpi is a well-known resolution measure that communicates how many pixels are required to represent one inch in the real-world, as for example, when hardcopy printing. If insufficient pixels are caught per inch of the spray card, it winds up difficult to estimate the tiniest drops' width. This may impact the diameter distribution analysis concealing issues in the spraying procedure.

To refine our conclusions, we experimented on the minimum dpi that is fundamental for each drop diameter. In Table~\ref{tab:dpis}, one can see the minimum number of pixels to express each drop diameter for each dpi value; see that a few cells of the table are vacant (loaded up with a hyphen) showing that the diameter cannot be computationally represented in that dpi resolution. Likewise, see that, in the columns, the number of pixels for one same diameter increments with the resolution. Clearly, the more data, the more accuracy at the expense of more processing power, considerable more storage, and more system transmission demands when transferring pictures. From the table, it is conceivable to reason that 600 dpi is the minimum resolution for robust analyses, since it can represent diameters as small as 50 $\mu m$; meanwhile, a resolution of 1,200 dpi, albeit more robust, might prompt downsides with respect to the administration of image files that are too huge. In any case, even if a resolution is sufficient to represent a given diameter, it is not assured that drops with that diameter will be detected; this comes from the fact that the image detection relies upon different factors, for example, the nature of the focal points, and the image processing algorithm.

Table~\ref{tab:dpis} is a guide for developers willing to computationally analyze spray cards, and furthermore for agronomists who are choosing which hardware to purchase in the face of their needs.

\begin{table*}[!htb]
   \centering
   \resizebox{.5\columnwidth}{!}{
   \begin{tabular}{c|c|c|c|c|c|c|c}
       \hline
       \backslashbox{$\mu m$\kern-1em}{\kern-1em dpi} & 50 & 100 & 300 & 600 & 1200 & 2400 & 2600 \\ \hline
       10     & -  & -   & -   & -   & -    & -    & 1    \\ \hline
       50     & -  & -   & -   & 1   & 2    & 5    & 5    \\ \hline
       100    & -  & -   & 1   & 2   & 5    & 9    & 10   \\ \hline
       250    & -  & 1   & 3   & 6   & 12   & 24   & 26   \\ \hline
       500    & 1  & 2   & 6   & 12  & 24   & 47   & 51   \\ \hline
       1,000  & 2  & 4   & 12  & 24  & 47   & 94   & 102  \\ \hline
       10,000 & 20 & 39  & 118 & 236 & 472  & 945  & 1024 \\ \hline
   \end{tabular}}
   \caption{Pixels needed to represent a given length, given a dpi.}
   \label{tab:dpis}
\end{table*}

\section{Conclusions}
\label{sec:conclusao}
We presented DropLeaf, a portable application to quantify pesticide spray coverage via image processing of water-sensitive spray cards.
We demonstrated that the accuracy of DropLeaf was sufficient to permit the utilization of mobile phones as substitutes for costly and cumbersome equipment. The approach was instantiated in a freely-accessible tool to be utilized in the assessment of real-world crops -- \url{https://play.google.com/store/apps/details?id=upvision.dropleaf}. We experimented with the tool with two datasets of water-sensitive papers; our investigations exhibited that DropLeaf track drops with high precision, producing standard metrics for quantifying the pesticide coverage. Moreover, our portable application identifies overlapping drops, a significant improvement with respect to former methods because, by providing a finer precision, the tool produces better accuracy and more information. DropLeaf can be used in a range of farming applications, including the evaluation of emerging innovations of agricultural sprayers based on Unmanned Aerial Vehicles.

\section*{Acknowledgements}
The authors are thankful to Dr. Claudia Carvalho, who kindly provided annotated cards.
This work was partially supported by the Coordena\c{c}\~ao de Aperfei\c{c}oamento de Pessoal de N\'ivel Superior -- Brazil (CAPES) -- Finance Code 001, through grant 167967/2017-7; Funda\c{c}\~ao de Amparo \`a Pesquisa do Estado de S\~ao Paulo (FAPESP), through grants 2014/25337-0, 2016/17078-0, 2017/08376-0, 2018/17620-5, 2019/04461-9, 2020/07200-9, and CEPID CeMEAI grant 2013/07375-0); Conselho Nacional de Desenvolvimento Cient\'ifico e Tecnol\'ogico (CNPq) through grants 406550/2018-2, 305580/2017-5, and 05580/2017-5.

\section*{References}
\bibliographystyle{elsarticle-harv}
\bibliography{references}
\end{document}